\documentclass[pdflatex,sn-mathphys-num]{sn-jnl}

\usepackage{graphicx}%
\usepackage{multirow}%
\usepackage{amsmath,amssymb,amsfonts}%
\usepackage{amsthm}%
\usepackage{mathrsfs}%
\usepackage[title]{appendix}%
\usepackage{xcolor}%
\usepackage{textcomp}%
\usepackage{manyfoot}%
\usepackage{booktabs}%
\usepackage{algorithm}%
\usepackage{algorithmicx}%
\usepackage{algpseudocode}%
\usepackage{listings}%
\usepackage{siunitx}
\usepackage{float}
\usepackage{comment}
\usepackage{lmodern}

\theoremstyle{thmstyleone}%
\theoremstyle{thmstyletwo}%

\theoremstyle{thmstylethree}%

\begin{document}

\title[DefSynUS: Real-time Patient-specific Intrahepatic Vessel Identification]{DefSynUS: Real-time Patient-specific Intrahepatic Vessel Identification via Deformation-Aware CT-US Domain Adaptation}


\author*[1,2]{\fnm{Karl-Philippe} \sur{Beaudet}}\email{karl-philippe.beaudet@inria.fr}
\author[3]{\fnm{Yordanka} \sur{Velikova}}
\author[1,4]{\fnm{Sidaty} \sur{El Hadramy}}
\author[3]{\fnm{Nassir} \sur{Navab}}
\author[4]{\fnm{Philippe} \sur{Cattin}}
\author[1,2,5]{\fnm{Juan} \sur{Verde}}
\author[1,2]{\fnm{Stéphane} \sur{Cotin}}

\affil[1]{\orgname{Inria}, \orgaddress{\city{Strasbourg}, \country{France}}}

\affil[2]{\orgname{University of Strasbourg}, \orgaddress{\city{Strasbourg}, \country{France}}}

\affil[3]{\orgname{Technical University of Munich}, \orgaddress{\country{Germany}}}

\affil[4]{\orgname{University of Basel}, \orgaddress{\country{Switzerland}}}

\affil[5]{\orgname{Institute of Image-Guided Surgery}, \orgaddress{\country{France}}}


\abstract{
\textbf{Purpose:} Laparoscopic ultrasound (LUS) enhances the safety of liver surgery by visualizing intrahepatic vessels in real-time. Still, vessel identification remains difficult due to probe constraints, complex vascular structure, and tissue deformation. This work aims to enable real-time, patient-specific vessel identification that remains robust under deformation through deformable ultrasound augmentation.  

\textbf{Methods:} Preoperative CT vessel annotations are used to generate synthetic ultrasound data via optimized physics-based rendering, coupled with domain adaptation to intraoperative ultrasound. The rendering is trained end-to-end for vessel identification and patient-specificity, eliminating the need for preoperative ultrasound. A deformation-aware augmentation simulates realistic intraoperative motion and tissue deformation within the rendering pipeline.

\textbf{Results:} In abdominal phantom and limited clinical feasibility experiments (single-case clinical evaluation), the framework achieved real-time intrahepatic vessel-branch identification, maintaining performance under new patient poses. 

\textbf{Conclusion:} The framework enables real-time vessel identification without preoperative ultrasound and supports technical feasibility, but multi-patient validation is still needed for generalizability and clinical feasibility.
}

\keywords{Ultrasound Imaging, Vessel Identification, Liver Surgery, Deformation Modeling, Data Augmentation}



\maketitle

\section{Introduction}

Liver resection is a commonly performed treatment that often offers the best chance for tumor removal and potential cure. The introduction of laparoscopic techniques has further advanced the treatment, demonstrating potential in reducing postoperative complications and improving patient outcomes \cite{fretland2018laparoscopic}. However, these procedures are highly complex and require precise imaging for optimal guidance. Intraoperative ultrasound (IOUS) has emerged as a valuable tool, allowing surgeons to evaluate tumor margins and prevent vascular damage \cite{ciria2016comparative}. It offers a radiation-free and non-invasive imaging method for real-time assessment during the intervention \cite{el2023trackerless,hagopian2014abdominal}. In laparoscopic surgery, using IOUS presents challenges due to limited visibility and confined space \cite{hagopian2014abdominal}. The small incisions and specialized instruments make it difficult to accurately position the laparoscopic IOUS (LIOUS) probe. Consequently, image interpretation becomes challenging, as it requires viewing ultrasound in a longitudinal-like plane, demanding the surgeon to relearn and retrain. This makes identifying intrahepatic vessels particularly difficult, as the limited range of motion within the abdominal cavity further complicates the process of obtaining clear and interpretable images \cite{beaudet2024towards}. Identifying intrahepatic vessels in ultrasound is essential for precise liver resection. During the procedure, the clinician must differentiate between the portal and hepatic veins and recognize key portal vein branches, such as the Main Portal Vein (MPV), Right Portal Vein (RPV), and Left Portal Vein (LPV). This distinction is crucial for spatial awareness and precise navigation within the liver.

Beyond limited field of view, soft-tissue deformation from respiration, pneumoperitoneum, manipulation, and probe pressure degrades guidance and confounds vessel recognition. Prior work addresses this with nonrigid US–preoperative registration \cite{smit2024ultrasound}, sparse-surface and vascular-feature–driven intraoperative deformation correction \cite{heiselman2020intraoperative}, model-based compensation evaluated with tracked US \cite{clements2016evaluation}, and respiration-induced fusion error correction \cite{yang2016ultrasound}. Recent learning-based 2D–3D registration further estimates deformation fields in real-time \cite{wang2025eureg}. Despite these advances, most approaches still depend on robust intraoperative registration between a preoperative volume and dedicated intraoperative US data.

Recent deep learning has enabled automatic segmentation of intrahepatic vessels in 2D \cite{Wei2019-xb}, typically with CNNs such as U\textendash Net \cite{ronneberger2015u} and, more recently, transformer-based models \cite{Wu2023-ac,Hille2024-sb} to capture long-range dependencies. These methods segment vascular trees accurately but do not identify vessel branches. Branch identification is difficult because vascular topology varies across patients and per-branch labeled ultrasound is scarce, which limits cross-patient generalization and often forces patient-specific training.

In laparoscopic liver surgery, intrahepatic vessel identification poses a dual challenge: clinically, surgeons need reliable real-time recognition of MPV/LPV/RPV and hepatic veins under constrained probe motion; technically, models must be patient-specific and deformation-aware. Patient-specific deep learning that leverages preoperative anatomy is promising. One strategy trains on 3D US volumes reconstructed from tracked 2D acquisitions for real-time guidance \cite{beaudet2024towards}, but it depends on tracked preoperative ultrasound (not routinely acquired) and overlooks intraoperative deformations. Preoperative CT, by contrast, is standard-of-care and provides high-resolution anatomy, offering a more practical foundation.

This work introduces a patient-specific vessel identification method using preoperative CT and unpaired ultrasound. To bridge the CT–US domain gap, we leverage recent CT-to-US domain adaptation. Among existing techniques \cite{velikova2023lotus,Velikova2024-fn}, physics-based rendering preserves spatial vessel relationships and avoids anatomically implausible hallucinations. Building on this, we adopt a differentiable simulation pipeline to generate ultrasound-like images from CT, enabling task-specific adaptation and real-time inference while maintaining high fidelity to vascular structures. Crucially, we integrate \emph{deformation-aware augmentation} into the physics-based rendering loop to mimic intraoperative motion and tissue deformation, improving robustness without requiring explicit intraoperative registration.

\textbf{Contributions.} (i) A patient-specific surgical framework trained solely from preoperative CT, eliminating the need for preoperative US acquisition; (ii) a deformation-aware, patient-specific reslicing augmentation module (new) integrated upstream of rendering, while reusing LOTUS/CUT as domain-adaptation backbones, to improve robustness to intraoperative deformation under CT-only patient supervision; and (iii) real-time intrahepatic vessel identification validated on an abdominal phantom and in a single-case clinical transcutaneous feasibility study, demonstrating technical feasibility under constrained ultrasound conditions. This approach is designed to enhance precision during minimally invasive liver resection without disrupting established clinical workflows. Given the limited clinical validation (single-case clinical study), this work is positioned as a feasibility study rather than a definitive demonstration of generalizability or clinical efficacy. The remainder of the paper is organized as follows: Section~\ref{sec:Methods} details the proposed method, Section~\ref{sec:Results} presents the experimental results, and Section~\ref{sec:conclusion} concludes with future directions.

\section{Methodology}\label{sec:Methods}

\begin{figure}[H]
    \centering
    \includegraphics[width=\linewidth]{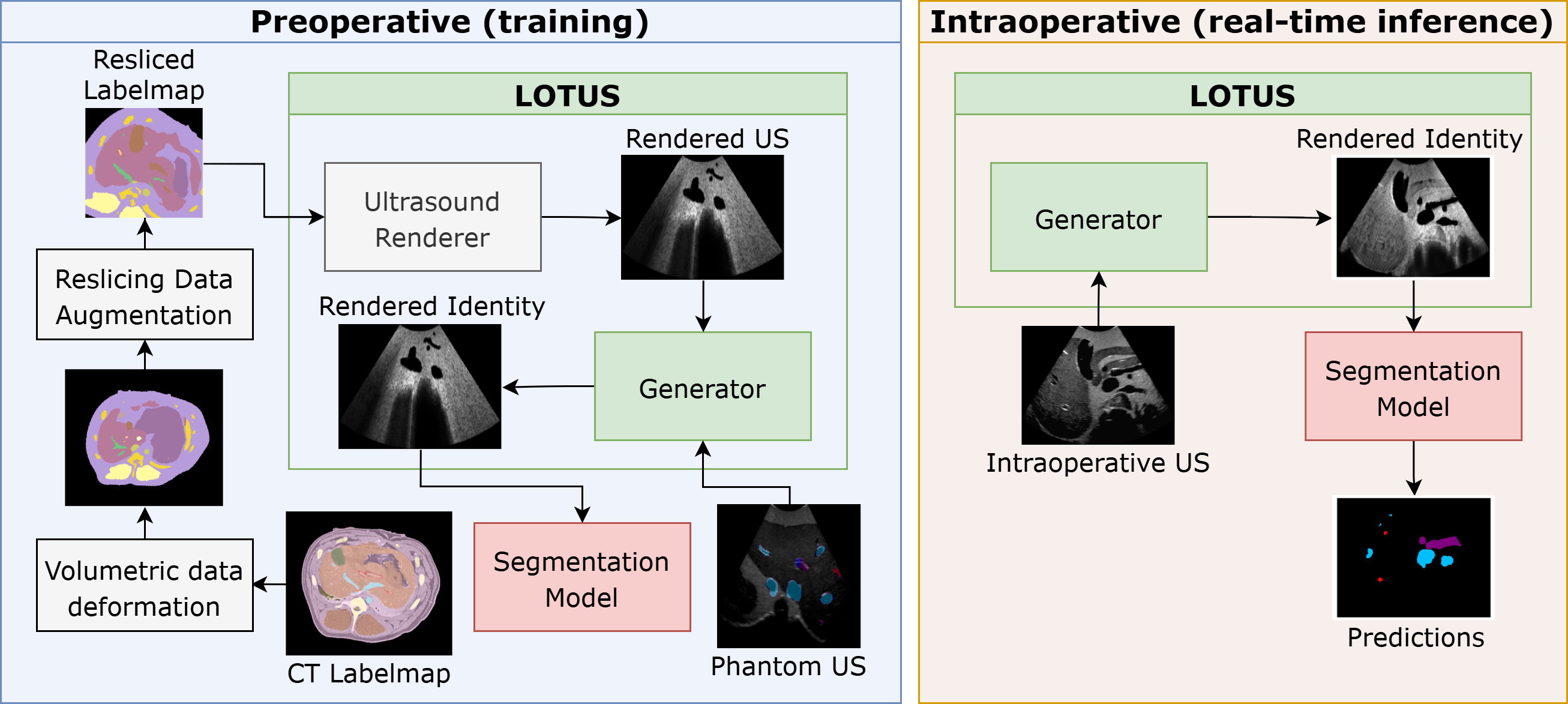}
    \caption{Overview of the proposed patient-specific method for intrahepatic vessel identification. A segmentation model is trained on ultrasound images simulated from the label map of the preoperative CT using physics-based rendering. During surgery, a trained image-to-image translation model bridges the gap between real and rendered ultrasound, enabling vessel identification on real intraoperative ultrasound.}
    \label{fig:method_overview}
\end{figure}

The proposed approach leverages preoperative CT data and domain adaptation to achieve patient-specific intrahepatic vessel identification on intraoperative US images. It consists of two main phases: (1) a preoperative training phase and (2) an intraoperative inference phase, as illustrated in Fig.~\ref{fig:method_overview}. 

DefSynUS does not require patient-specific preoperative ultrasound acquisition. The vessel identification model is trained solely from the patient’s CT-derived vessel labels (in the proposed workflow, this can be run at least one day before surgery). In configurations using patient ultrasound for domain adaptation, real ultrasound is used only to train the image translation model (CUT), not to supervise vessel labels, in the CT-only configuration, even this adaptation step uses phantom ultrasound only.

\subsection{Preoperative training phase}

\subsubsection{Data preparation and augmentation}
We begin with a CT label map containing annotated intrahepatic vessels from the patient's preoperative CT scan. 

\noindent\textbf{Deformation-aware augmentation.}
To be robust to organ motion and/or deformation that occurs between CT acquisition and US-based vessel identification, we augment our dataset with volumetric deformations. Let $L_{\text{CT}}:\Omega\subset\mathbb{R}^3\!\to\!\{0,\ldots,K\}$ denote the voxelwise label map and $I_{\text{CT}}:\Omega\!\to\!\mathbb{R}$ the corresponding CT. 
Rather than modeling explicit biomechanics, we sample a broad but controlled distribution of smooth 3D displacement fields $u:\Omega\!\to\!\mathbb{R}^3$ to train stability under plausible intraoperative deformations.
Each $u$ is generated by drawing a coarse random vector field on a low-resolution lattice and upsampling it to the region of interest with trilinear interpolation, followed by Gaussian smoothing and amplitude scaling.
We constrain per-axis displacement to $[-42.63,\,41.05]\,\mathrm{mm}$ (mean absolute amplitude $26.65\,\mathrm{mm}$ per axis) to control the deformation distribution and prevent unrealistic warps.
The warped label map and CT are then
\begin{equation}
    \tilde{L}_{\text{CT}}(x) \!=\! L_{\text{CT}}\!\big(x + u(x)\big),\quad
    \tilde{I}_{\text{CT}}(x) \!=\! I_{\text{CT}}\!\big(x + u(x)\big),
\end{equation}
implemented via grid sampling with nearest-neighbor interpolation for $\tilde{L}_{\text{CT}}$ and (optionally) bilinear for $\tilde{I}_{\text{CT}}$. To ensure samples remain within the domain, we apply padding before the deformation as described in \cite{lecomte2025domain}. 
This step yields a bank of plausibly deformed anatomies per patient without requiring explicit registration or motion models. This \emph{deformation-aware augmentation} is performed before any 2D reslicing.

\noindent\textbf{Label map reslicing.}
After deformation, we apply a reslicing strategy similar to~\cite{beaudet2024towards}. The deformed label map $\tilde{L}_{\text{CT}}$ is resliced along multiple planes to simulate diverse LIOUS and transcutaneous views, including tilt, rock, slide, and small out-of-plane rotations. Formally, for a given slicing transform $T:\mathbb{R}^2\!\to\!\mathbb{R}^3$,
\begin{equation}
    L(x,y) = \mathcal{S}\big\{\tilde{L}_{\text{CT}}\!\big(T(x,y)\big)\big\},
    \label{eq:reslicing}
\end{equation}
where $\mathcal{S}$ denotes spline interpolation on the 3D grid. By reslicing \emph{after} deformation, each 2D slice encodes realistic in-plane distortions and topology-preserving but nonrigid changes in vessel cross-sections. This produces a dataset of 2D label maps that captures patient-specific branching patterns across clinically plausible probe orientations.

\noindent\textbf{Ultrasound simulation.}
We employ the LOTUS framework~\cite{velikova2023lotus} for physics-based ultrasound simulation. It consists of two primary components:
\begin{itemize}
    \item \textbf{Ultrasound renderer.} The resliced 2D label maps are fed into the differentiable renderer, where each tissue class is parameterized by attenuation ($\alpha$), acoustic impedance ($Z$), and three speckle parameters ($\mu_0,\mu_1,\sigma_0$)~\cite{velikova2023lotus}. The renderer composes attenuation, reflection, and scatter sub-maps and traces scanlines to mimic acoustic propagation.
    \item \textbf{Generator.} A lightweight generator refines rendered images toward the appearance of intraoperative ultrasound, narrowing the synthetic–real gap while preserving anatomy.
\end{itemize}
The differentiability of the renderer enables end-to-end optimization with task losses. During training, rendering parameters and the generator adapt to produce US representations that are maximally informative for vessel identification, tailored to each patient’s anatomy and the deformation distribution induced above.

\subsection{Patient-specific training:}
Using the paired data (simulated ultrasound images and their corresponding vessel annotations), we train a Segmentation Model based on the Attention U-Net architecture as described in \cite{beaudet2024towards}, which has shown robust performance in intrahepatic vessel identification. This model is specifically tailored to identify and segment the intrahepatic vessels of the individual patient. 
To bridge the appearance gap between rendered and real ultrasound images, we incorporate an unpaired image-to-image translation network based on the CUT framework \cite{park2020contrastive}. The generator $G$ learns a mapping $G: X \rightarrow Y$ from the source domain $X$ (real ultrasound) to the target domain $Y$ (rendered ultrasound) using the following adversarial loss:
\begin{equation}
    L_{\text{GAN}} = \mathbb{E}_{y\sim Y}\left[\log D(y)\right] + \mathbb{E}_{x\sim X}\left[\log\bigl(1-D(G(x))\bigr)\right],
\end{equation}
where $D$ is the discriminator network. In addition, a contrastive loss is applied on sampled image patches to ensure that anatomical details are preserved during translation. Note that we choose to learn the mapping to the synthetic image as this task is easier than translation towards a real US image.

The total loss function for our framework combines segmentation accuracy and domain alignment objectives:

\begin{equation}
\mathcal{L}_{\text{total}} = \lambda_{\text{seg}}\mathcal{L}_{\text{dice}} + \lambda_{\text{domain}}\mathcal{L}_{\text{GAN}} + \lambda_{\text{contrastive}}\mathcal{L}_{\text{contrastive}}
\end{equation}
where $\mathcal{L}_{\text{dice}}$ is the segmentation loss between predicted segmentation and ground truth label map, $\mathcal{L}_{\text{GAN}}$ is the adversarial loss, and $\mathcal{L}_{\text{contrastive}}$ preserves anatomical features. The hyperparameters balance the contribution of each loss term.

\subsection{Real-time intraoperative identification}
During the intervention, the ultrasound probe continuously captures real-time images. These images are first processed by the \textbf{domain adaptation network}, which aligns them with the domain used for training the segmentation model. The adapted images are then fed into the \textbf{segmentation model} (Attention U-Net), which performs real-time identification of key intrahepatic vascular branches, including the MPV, LPV, RPV, and HV. This real-time pipeline ensures fast and accurate vessel identification, making it well-suited for intraoperative use.

We assume the CT$\rightarrow$US pipeline preserves vascular topology: the renderer maintains geometry and the mapping $G$ (real$\to$rendered) alters appearance without anatomical distortions. Deformations are smooth, bounded fields capturing respiration, pneumoperitoneum, probe pressure, and manipulation; we do not model sliding along fissures, local vessel collapse, highly heterogeneous strain, or topology changes.

\section{Results}\label{sec:Results}

\subsection{Dataset}

We use two cohorts with distinct roles: (i) a CT/US-compatible torso phantom for controlled evaluation and as an external US source to train CUT in the \emph{CT-only} setting; and (ii) a tracked clinical transcutaneous cohort for cross-deformation evaluation.

\noindent\textbf{Phantom cohort (two roles).}
We acquired a CT scan and tracked transcutaneous US from a non-deformable CT/US-compatible torso phantom. The CT was segmented with TotalSegmentator~\cite{wasserthal2023totalsegmentator} and refined to label MPV, RPV, LPV, and HV. From this label map, we generated 25{,}000 2D slices using reslicing (Eq.~\eqref{eq:reslicing}) without deformation-aware augmentation, since the phantom anatomy is rigid, to train the vessel identification model via physics-based rendering. Tracked US was acquired with a 5C1 probe (ACUSON Juniper, Siemens) mounted on a KUKA LBR iiwa, enabling pose estimation and CT-aligned labeling.
Phantom \emph{real} US is used (1) for phantom evaluation (held-out test set) and (2) to train CUT (real$\rightarrow$rendered) in the \emph{CT-only} setting, without any patient ultrasound.
The phantom tracked US comprises 333 frames: 272 for CUT training, 7 for CUT validation/early stopping, and 54 for testing (phantom evaluation).

\noindent\textbf{Clinical data.}
Clinical tracked transcutaneous US was acquired with an ACUSON S3000 (6C1 probe, 13\,cm depth) and EM tracking (trakSTAR, NDI). Video was captured via frame grabber (AV.io HD+, Epiphan) and recorded with PLUS/3D Slicer; spatial and temporal calibration used fCal (PLUS). Frames are $680\times 480$ pixels with 0.3\,mm spacing.

The clinical cohort is a single healthy adult (feasibility study) with paired contrast-enhanced CT and tracked transcutaneous US. Inclusion required an abdominal/vessel label map, adequate US quality, and successful tracking/calibration; exclusion criteria (none met) included prior hepatic surgery, diffuse liver disease (e.g., cirrhosis), and motion-corrupted tracking. The cohort comprises 876 tracked US frames: 718 for CUT training, 16 for CUT validation, and 142 for testing.
For patient-specific training in the clinical experiments, the clinical CT label map was augmented with deformation-aware augmentation (Sec.~\ref{sec:Methods}) followed by reslicing (Eq.~\eqref{eq:reslicing}) to generate 25{,}000 simulated training slices.

\subsection{Implementation}
The proposed framework is implemented in Python using PyTorch 2.5.1 \footnote{\url{https://pytorch.org/docs/stable/index.html}} and MONAI 1.3.0 \footnote{\url{https://docs.monai.io/en/stable/index.html}}. The implementation of the differentiable ultrasound renderer is adapted from LOTUS \cite{velikova2023lotus}. The reslicing augmentation is performed with MONAI's \texttt{RandAffine} transform, which applies random affine transformations (including rotation, tilting, rocking, and sliding). 
The Attention U-Net \cite{Oktay2018AttentionUNet} (MONAI implementation) is trained using the Adam optimizer with a learning rate of $1 \times 10^{-5}$ and a batch size of $8$. Models are trained for $10$–$200$ epochs with early stopping, terminating training when the stopping criterion fails to decrease for at least $5$ consecutive epochs. 
For the LOTUS framework, we configure the ultrasound renderer to simulate a curvilinear ultrasound probe. The tissue parameters (attenuation coefficient, acoustic impedance, and speckle distribution parameters) are initialized based on literature values \cite{velikova2023lotus} and then optimized during training.
The adversarial generator and discriminator follow the architecture described in the CUT framework \cite{park2020contrastive}, with PatchGAN discriminators and a ResNet-based generator. Training for a single patient completes in $\sim$2 hours; intraoperative inference runs at $56.6$ fps on an RTX 4070~Ti.

\subsection{Evaluation Metrics} \label{sec:Metric}

We report precision and recall for vessel identification and branch classification. A \emph{predicted island} is a connected component of predicted vessel pixels with area $\geq$50\,px. Matching is performed at the island level: a predicted island is a true positive (TP) if there exists a ground-truth (GT) island of the same label whose minimum surface-to-surface distance is $\leq$5\,mm; otherwise it is a false positive (FP). Unmatched GT islands are false negatives (FN). True negatives (TN) are not defined in this setting since we only consider precision and recall. The 5\,mm tolerance, adopted from \cite{beaudet2024towards}, is chosen \emph{a priori} to absorb residual tracking/calibration error, reslicing imprecision, and small uncompensated deformations.

\subsection{Experiments and Results}  

We evaluated hepatic vessel-branch classification (MPV, LPV, RPV, HV), comparing CT-to-US domain adaptation methods against a baseline trained on preoperative ultrasound. Five configurations were considered: (i) \textbf{Baseline} \cite{beaudet2024towards}; (ii) \textbf{LOTUS} \cite{velikova2023lotus}; (iii) \textbf{DefSynUS-rigid} i.e. LOTUS with realistic ultrasound reslicing augmentation; (iv) \textbf{DefSynUS-Unet}, i.e. LOTUS with deformation-aware reslicing augmentation; and (v) \textbf{DefSynUS} (ours), i.e. LOTUS with advanced deformation-aware augmentation and an Attention U-Net head. 

We further evaluated a CT-only scenario in which \textbf{no patient-specific preoperative ultrasound was used at any stage}. The vessel identification model was trained exclusively from the patient’s CT-derived data; abdominal phantom ultrasound was used only for CUT training and stopping-criterion selection.

We designed two complementary experiments to probe generalization across acquisition settings and deformation. First, a \textbf{phantom transcutaneous ultrasound} where CT-to-US domain adaptation was performed between CT-derived renderings and real transcutaneous US acquired on a CT/US-compatible phantom. Second, a \textbf{clinical transcutaneous ultrasound}, where CT-to-US domain adaptation was evaluated on clinical transcutaneous US sequences while explicitly accounting for deformation factors (probe compression and respiratory phase).

\subsubsection{Phantom validation}

\begin{figure}[htbp]
    \centering
    \includegraphics[width=\linewidth]{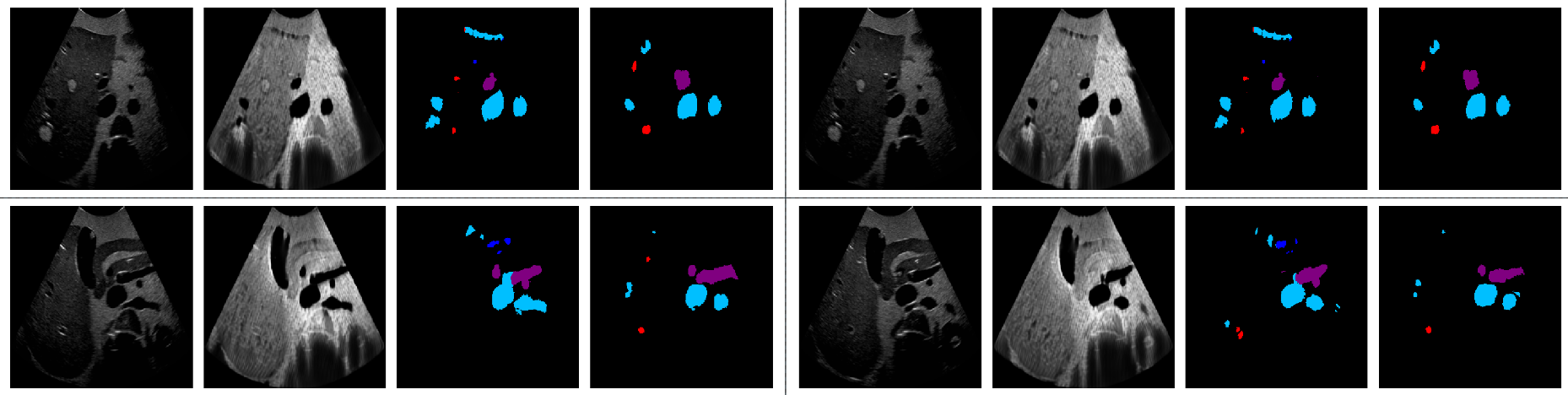}
    \caption{Vessel classification of MPV, LPV, RPV, and HV. Four prediction examples are shown. From left to right: (i) Real Ultrasound, (ii) rendered ultrasound identity, (iii) segmentation model prediction, (iv) vessel classification label map (ground truth). Examples shown from experiments where all four classes are available.}
    \label{fig:predictions}
\end{figure}

Table~\ref{tab:unet_classification} reports precision and recall per class and averaged across classes, and Fig.~\ref{fig:predictions} provides qualitative examples.
U\mbox{-}Net~LOTUS performed well for MPV and HV but struggled with LPV, resulting in a mean precision of $0.50 \pm 0.28$.
DefSynUS-rigid improved MPV precision and increased mean precision ($0.53 \pm 0.24$), but LPV remained challenging. Importantly, the observed differences between methods are small relative to the frame-level variability, with substantial overlap in precision distributions across configurations. In the phantom experiment, we report only rigid/reslicing-based configurations (e.g., DefSynUS-rigid) and do not include deformation-aware augmentation, because the phantom anatomy is non-deformable and does not exhibit the tissue motion/compression effects that the proposed deformation-aware augmentation is designed to model.

The zero precision and recall values observed for LPV in the Baseline are primarily explained by (i) class imbalance and (ii) anatomical difficulty. In this transcutaneous phantom acquisition, LPV instances are less frequent and typically smaller in caliber than MPV or HV, producing fewer and smaller islands that are more sensitive to partial-volume effects and speckle. Under our island-level metric, these missed detections dominate the score, yielding zero values despite occasional correct localization in individual frames.

Note that in this transcutaneous validation, RPV was not evaluated because the scan targeted the left liver, limiting the field of view to MPV, LPV, and HV. This setup does not fully reflect intraoperative imaging, as all ultrasound images were acquired near the axial plane and do not capture the full variability of probe orientation typically encountered during surgery. Importantly, the goal here is not to outperform the baseline but to \emph{achieve comparable results without relying on preoperative ultrasound}, validating the feasibility of CT-to-US domain adaptation. In particular, these experiments indicate that \emph{domain adaptation combined with reslicing-based augmentation can learn hepatic vessel appearance in ultrasound} from preoperative CT-derived label maps.
Accordingly, this experiment assesses feasibility rather than statistical superiority, evaluating whether CT-to-US domain adaptation can achieve stable vessel identification under realistic variability. The results indicate that domain adaptation combined with reslicing-based augmentation can learn hepatic vessel appearance in ultrasound from preoperative CT-derived label maps.

\begin{table}[ht]
    \centering
    \caption{Precision and recall for hepatic vessel-branch classification in the transcutaneous ultrasound phantom experiment}
    \label{tab:unet_classification}
    \begin{tabular*}{\textwidth}{@{\extracolsep\fill}lcccc}
        \toprule
        Model & MPV & LPV & HV & \textbf{Mean} \\
        \midrule
        \multicolumn{5}{c}{\textbf{Precision}} \\
        \midrule
        Baseline \cite{beaudet2024towards} & $0.66 \pm 0.47$ & $0.00 \pm 0.00$ & $\mathbf{0.95 \pm 0.14}$ & $\mathbf{0.63 \pm 0.46}$ \\
        \midrule
        LOTUS \cite{velikova2023lotus} & $0.65 \pm 0.46$ & $\mathbf{0.25 \pm 0.43}$ & $0.67 \pm 0.23$ & $0.50 \pm 0.28$ \\
        DefSynUS-rigid & $\mathbf{0.84 \pm 0.29}$ & $0.12 \pm 0.31$ & $0.63 \pm 0.24$ & $0.53 \pm 0.24$ \\
        \midrule
        \multicolumn{5}{c}{\textbf{Recall}} \\
        \midrule
        Baseline \cite{beaudet2024towards} & $0.58 \pm 0.45$ & $0.00 \pm 0.00$ & $0.56 \pm 0.22$ & $0.55 \pm 0.44$ \\
        \midrule
        LOTUS \cite{velikova2023lotus} & $0.59 \pm 0.45$ & $0.25 \pm 0.43$ & $0.63 \pm 0.24$ & $0.46 \pm 0.27$ \\
        DefSynUS-rigid & $0.84 \pm 0.29$ & $0.14 \pm 0.35$ & $0.66 \pm 0.20$ & $0.55 \pm 0.21$ \\
        \bottomrule
    \end{tabular*}
\end{table}

\subsubsection{Clinical cross-deformation validation}

We evaluated cross\mbox{-}deformation generalization on clinical data using preoperative CT label maps and a \emph{different\mbox{-}pose} transcutaneous US sequence with altered probe compression and apnea cycle. Models were trained with CT\mbox{-}to\mbox{-}US domain adaptation using transcutaneous US from the reference pose (Def$_1$) and evaluated on the distinct pose (Def$_2$) without pose\mbox{-}specific fine\mbox{-}tuning. Mean precision and recall were computed using the island\mbox{-}level matching protocol in Sec.~\ref{sec:Metric} (Table~\ref{tab:Human_IOUS}). We compared: (i) \textbf{Baseline} \cite{beaudet2024towards} on the \emph{same} deformation (Def$_1\!\rightarrow\!$Def$_1$) and a \emph{different} deformation (Def$_1\!\rightarrow\!$Def$_2$); (ii) \textbf{LOTUS} \cite{velikova2023lotus} trained on Def$_1$ and tested on Def$_2$; (iii) \textbf{DefSynUS-Unet}, which adds deformation\mbox{-}aware reslicing augmentation; and (iv) \textbf{DefSynUS} (ours), which further integrates advanced deformation\mbox{-}aware augmentation and an Attention U\mbox{-}Net head.

In addition to these configurations, we evaluated (v) a clinically realistic \emph{CT-only} setting in which \textbf{no patient preoperative ultrasound was used at any stage}. In this configuration, the intraoperative vessel identification model was trained exclusively from the patient’s CT-derived data, while abdominal phantom ultrasound was used solely to train the CUT model and to define the stopping criterion.

As summarized in Table~\ref{tab:Human_IOUS}, \textbf{DefSynUS} attains the best balance under pose and compression changes ($0.46\!\pm\!0.33$ precision, $0.48\!\pm\!0.38$ recall), outperforming LOTUS and the cross\mbox{-}deformation Baseline. Notably, the CT-only configuration preserves this trend, yielding performance in the same range of the US-trained DefSynUS variant, indicating that deformation robustness, in this feasibility setting, is primarily driven by CT-derived augmentation rather than access to patient ultrasound. The improvement of \textbf{DefSynUS-Unet} over LOTUS indicates that deformation\mbox{-}targeted augmentation contributes to robustness, while the full configuration yields additional gains, suggesting that attention\mbox{-}based segmentation plus stronger augmentation jointly reduce false\mbox{-}positive branch assignments. 



\begin{table}[ht]
    \centering
    \caption{Mean Precision and recall for hepatic vessel-branch classification in the clinical cross-deformation experiment}
    \label{tab:Human_IOUS}
    \begin{tabular*}{0.8\textwidth}{@{\extracolsep\fill}lcc}
        \toprule
        Model & Precision & Recall \\
        \midrule
        Baseline (Def$_1\!\rightarrow\!$Def$_1$) & $0.43 \pm 0.38$ & $0.25 \pm 0.31$ \\
        Baseline (Def$_1\!\rightarrow\!$Def$_2$) & $0.33 \pm 0.32$ & $0.22 \pm 0.29$ \\
        \midrule
        LOTUS \cite{velikova2023lotus} & $0.35 \pm 0.30$ & $0.28 \pm 0.32$ \\
        DefSynUS-Unet & $0.43 \pm 0.40$ & $0.34 \pm 0.35$ \\
        DefSynUS (With preOP US) & $0.46 \pm 0.33$ & $\mathbf{0.48 \pm 0.38}$ \\
        \textbf{DefSynUS (Ours, without preOP US)} & $\mathbf{0.48 \pm 0.34}$ & $0.46 \pm 0.36$ \\
        \bottomrule
    \end{tabular*}
\end{table}

\section{Discussion}\label{sec:discussion}

DefSynUS enables patient-specific intrahepatic vessel-branch identification (MPV/LPV/RPV/HV) by using routine preoperative CT for supervision and domain adaptation to operate on intraoperative ultrasound under deformation and probe variability.

\textbf{Phantom validation}
The phantom experiment serves as a controlled sanity check for the full pipeline in a setting where tracked, label-consistent ultrasound is feasible to acquire, with stable anatomy and no physiological deformation. It is not meant to replace clinical validation, but to verify the CT-to-US adaptation and reslicing-based training strategy before testing on clinical data.

\textbf{CT-only workflow.}
In the CT-only configuration, no patient ultrasound is used at any stage: the vessel identification model is trained exclusively from the patient’s CT-derived anatomy, while phantom ultrasound is used only to train CUT and define the stopping criterion. This separation maintains patient specificity while using external ultrasound solely to learn generic appearance statistics for translation.

\textbf{Precision-first and clinical risk.}
We prioritize precision over recall because false-positive vessel overlays can mislead intraoperative decision-making, whereas missed detections can often be mitigated using temporal context and probe repositioning. This conservative behavior supports use as an assistive overlay rather than an autonomous decision tool. Consistently, we assess performance with precision-centered metrics that better reflect clinical risk: the achieved precision falls within the range reported for surgeons in IOUS (0.05--0.54)~\cite{beaudet2024towards}. We also use a 5,mm island-matching tolerance to account for residual uncertainty from tracking/calibration and small uncompensated deformations, making the evaluation closer to what is clinically interpretable rather than an artificially strict pixel-level comparison.


\textbf{Anatomical variability and pathology.}
Patient-specific CT supervision directly captures individual vascular topology, addressing substantial inter-patient anatomical variability. However, our single healthy-liver case does not represent challenging conditions (e.g., cirrhosis, prior surgery, vascular remodeling), which may alter ultrasound appearance and vessel visibility; these cases should be included in future multi-patient prospective IOUS studies to better characterize robustness.

\textbf{Limitations and future work.}
Clinical results are currently limited to a single patient and transcutaneous US as a proxy for LIOUS; therefore, the present study should be interpreted as a technical/workflow feasibility study and not as evidence of broad generalizability, clinical efficacy, or deployment readiness. Rigorous quantitative validation is also constrained by the limited availability of tracked ultrasound acquired over many probe poses with accurately registered CT volumes and per-branch label maps, data that are costly and technically demanding to acquire and therefore uncommon in routine workflows. Future work will expand prospective IOUS validation, replace phantom ultrasound for translation with multi-patient ultrasound to improve generalization, and integrate uncertainty-aware visualization to suppress low-confidence overlays.




\section{Conclusion}\label{sec:conclusion}

We introduced DefSynUS, a real-time patient-specific framework for intrahepatic vessel-branch identification that trains from preoperative CT-derived labels and operates on intraoperative ultrasound through CT-to-US domain adaptation.
Across phantom and single-case clinical transcutaneous feasibility experiments, DefSynUS improved robustness to pose and compression changes and achieved stable real-time performance for MPV/LPV/RPV/HV identification. 
These results support the technical feasibility of CT-driven, patient-specific vessel identification in ultrasound-guided liver surgery, but broader multi-patient prospective IOUS validation is required to assess generalizability and clinical feasibility, and to support future integration with confidence-aware visualization for intraoperative use.

\section*{Declarations}

\begin{itemize}
\item Funding: No funding was received for this work.
\item Conflict of interest: The authors have no competing interests to declare that are relevant to the content of this article.
\item Ethics approval and consent to participate: This single-case retrospective technical validation on de-identified data was conducted in accordance with institutional policies.
\item Consent for publication: Not applicable
\item Data, code and/or material availability: Upon publication, the dataset and source code will be released at \url{https://github.com/Karl-Philippe/DefSynUS}. Licenses: data CC BY 4.0; code Apache-2.0.
\item Author contribution:
Karl-Philippe Beaudet: Methodology, Software, Validation, Investigation, Writing - original draft; Yordanka Velikova: Methodology, Writing - review \& editing; Sidaty El Hadramy: Conceptualization, Writing - review \& editing; Philippe C. Cattin: Supervision, Writing - review \& editing; Nassir Navab: Supervision; Juan Verde: Investigation, Data curation, Ressources, Validation, Supervision; Stéphane Cotin: Supervision, Writing -review \& editing.
\end{itemize}


\bigskip


\bibliography{sn-bibliography}


\begin{thebibliography}{20}
\ifx \bisbn   \undefined \def \bisbn  #1{ISBN #1}\fi
\ifx \binits  \undefined \def \binits#1{#1}\fi
\ifx \bauthor  \undefined \def \bauthor#1{#1}\fi
\ifx \batitle  \undefined \def \batitle#1{#1}\fi
\ifx \bjtitle  \undefined \def \bjtitle#1{#1}\fi
\ifx \bvolume  \undefined \def \bvolume#1{\textbf{#1}}\fi
\ifx \byear  \undefined \def \byear#1{#1}\fi
\ifx \bissue  \undefined \def \bissue#1{#1}\fi
\ifx \bfpage  \undefined \def \bfpage#1{#1}\fi
\ifx \blpage  \undefined \def \blpage #1{#1}\fi
\ifx \burl  \undefined \def \burl#1{\textsf{#1}}\fi
\ifx \doiurl  \undefined \def \doiurl#1{\url{https://doi.org/#1}}\fi
\ifx \betal  \undefined \def \betal{\textit{et al.}}\fi
\ifx \binstitute  \undefined \def \binstitute#1{#1}\fi
\ifx \binstitutionaled  \undefined \def \binstitutionaled#1{#1}\fi
\ifx \bctitle  \undefined \def \bctitle#1{#1}\fi
\ifx \beditor  \undefined \def \beditor#1{#1}\fi
\ifx \bpublisher  \undefined \def \bpublisher#1{#1}\fi
\ifx \bbtitle  \undefined \def \bbtitle#1{#1}\fi
\ifx \bedition  \undefined \def \bedition#1{#1}\fi
\ifx \bseriesno  \undefined \def \bseriesno#1{#1}\fi
\ifx \blocation  \undefined \def \blocation#1{#1}\fi
\ifx \bsertitle  \undefined \def \bsertitle#1{#1}\fi
\ifx \bsnm \undefined \def \bsnm#1{#1}\fi
\ifx \bsuffix \undefined \def \bsuffix#1{#1}\fi
\ifx \bparticle \undefined \def \bparticle#1{#1}\fi
\ifx \barticle \undefined \def \barticle#1{#1}\fi
\bibcommenthead
\ifx \bconfdate \undefined \def \bconfdate #1{#1}\fi
\ifx \botherref \undefined \def \botherref #1{#1}\fi
\ifx \url \undefined \def \url#1{\textsf{#1}}\fi
\ifx \bchapter \undefined \def \bchapter#1{#1}\fi
\ifx \bbook \undefined \def \bbook#1{#1}\fi
\ifx \bcomment \undefined \def \bcomment#1{#1}\fi
\ifx \oauthor \undefined \def \oauthor#1{#1}\fi
\ifx \citeauthoryear \undefined \def \citeauthoryear#1{#1}\fi
\ifx \endbibitem  \undefined \def \endbibitem {}\fi
\ifx \bconflocation  \undefined \def \bconflocation#1{#1}\fi
\ifx \arxivurl  \undefined \def \arxivurl#1{\textsf{#1}}\fi
\csname PreBibitemsHook\endcsname

\bibitem[\protect\citeauthoryear{Fretland et~al.}{2018}]{fretland2018laparoscopic}
\begin{barticle}
\bauthor{\bsnm{Fretland}, \binits{{\AA}.A.}},
\bauthor{\bsnm{Dagenborg}, \binits{V.J.}},
\bauthor{\bsnm{Bj{\o}rnelv}, \binits{G.M.W.}},
\bauthor{\bsnm{Kazaryan}, \binits{A.M.}},
\bauthor{\bsnm{Kristiansen}, \binits{R.}},
\bauthor{\bsnm{Fagerland}, \binits{M.W.}},
\bauthor{\bsnm{Hausken}, \binits{J.}},
\bauthor{\bsnm{T{\o}nnessen}, \binits{T.I.}},
\bauthor{\bsnm{Abildgaard}, \binits{A.}},
\bauthor{\bsnm{Barkhatov}, \binits{L.}},
\bauthor{\bsnm{Yaqub}, \binits{S.}},
\bauthor{\bsnm{R{\o}sok}, \binits{B.I.}},
\bauthor{\bsnm{Bj{\o}rnbeth}, \binits{B.A.}},
\bauthor{\bsnm{Andersen}, \binits{M.H.}},
\bauthor{\bsnm{Flatmark}, \binits{K.}},
\bauthor{\bsnm{Aas}, \binits{E.}},
\bauthor{\bsnm{Edwin}, \binits{B.}}:
\batitle{Laparoscopic {{Versus Open Resection}} for {{Colorectal Liver Metastases}}: {{The OSLO-COMET Randomized Controlled Trial}}}.
\bjtitle{Annals of Surgery}
\bvolume{267}(\bissue{2}),
\bfpage{199}--\blpage{207}
(\byear{2018})
\doiurl{10.1097/SLA.0000000000002353}
\end{barticle}
\endbibitem

\bibitem[\protect\citeauthoryear{Ciria et~al.}{2016}]{ciria2016comparative}
\begin{barticle}
\bauthor{\bsnm{Ciria}, \binits{R.}},
\bauthor{\bsnm{Cherqui}, \binits{D.}},
\bauthor{\bsnm{Geller}, \binits{D.A.}},
\bauthor{\bsnm{Briceno}, \binits{J.}},
\bauthor{\bsnm{Wakabayashi}, \binits{G.}}:
\batitle{Comparative {{Short-term Benefits}} of {{Laparoscopic Liver Resection}}: 9000 {{Cases}} and {{Climbing}}}.
\bjtitle{Annals of Surgery}
\bvolume{263}(\bissue{4}),
\bfpage{761}--\blpage{777}
(\byear{2016})
\doiurl{10.1097/SLA.0000000000001413}
\end{barticle}
\endbibitem

\bibitem[\protect\citeauthoryear{El~Hadramy et~al.}{2023}]{el2023trackerless}
\begin{bchapter}
\bauthor{\bsnm{El~Hadramy}, \binits{S.}},
\bauthor{\bsnm{Verde}, \binits{J.}},
\bauthor{\bsnm{Beaudet}, \binits{K.-P.}},
\bauthor{\bsnm{Padoy}, \binits{N.}},
\bauthor{\bsnm{Cotin}, \binits{S.}}:
\bctitle{Trackerless volume reconstruction from intraoperative ultrasound images}.
In: \bbtitle{Medical Image Computing and Computer-Assisted Intervention -- MICCAI 2023},
pp. \bfpage{303}--\blpage{312}.
\bpublisher{Springer},
\blocation{Cham}
(\byear{2023}).
\doiurl{10.1007/978-3-031-43999-5_29}
\end{bchapter}
\endbibitem

\bibitem[\protect\citeauthoryear{Hagopian and Machi}{2014}]{hagopian2014abdominal}
\begin{bbook}
\beditor{\bsnm{Hagopian}, \binits{E.J.}},
\beditor{\bsnm{Machi}, \binits{J.}} (eds.):
\bbtitle{Abdominal {{Ultrasound}} for {{Surgeons}}}.
\bpublisher{{Springer New York}},
\blocation{{New York, NY}}
(\byear{2014}).
\doiurl{10.1007/978-1-4614-9599-4}
\end{bbook}
\endbibitem

\bibitem[\protect\citeauthoryear{Beaudet et~al.}{2024}]{beaudet2024towards}
\begin{bchapter}
\bauthor{\bsnm{Beaudet}, \binits{K.-P.}},
\bauthor{\bsnm{Karargyris}, \binits{A.}},
\bauthor{\bsnm{El~Hadramy}, \binits{S.}},
\bauthor{\bsnm{Cotin}, \binits{S.}},
\bauthor{\bsnm{Mazellier}, \binits{J.-P.}},
\bauthor{\bsnm{Padoy}, \binits{N.}},
\bauthor{\bsnm{Verde}, \binits{J.}}:
\bctitle{Towards real-time intrahepatic vessel identification in intraoperative ultrasound-guided liver surgery}.
In: \bbtitle{Medical Image Computing and Computer-Assisted Intervention -- MICCAI 2024},
pp. \bfpage{649}--\blpage{659}.
\bpublisher{Springer},
\blocation{Cham}
(\byear{2024}).
\doiurl{10.1007/978-3-031-72089-5_61}
\end{bchapter}
\endbibitem

\bibitem[\protect\citeauthoryear{Smit et~al.}{2024}]{smit2024ultrasound}
\begin{barticle}
\bauthor{\bsnm{Smit}, \binits{J.N.}},
\bauthor{\bsnm{Kuhlmann}, \binits{K.F.D.}},
\bauthor{\bsnm{Thomson}, \binits{B.R.}},
\bauthor{\bsnm{Kok}, \binits{N.F.M.}},
\bauthor{\bsnm{Ruers}, \binits{T.J.M.}},
\bauthor{\bsnm{Fusaglia}, \binits{M.}}:
\batitle{Ultrasound guidance in navigated liver surgery: Toward deep-learning enhanced compensation of deformation and organ motion}.
\bjtitle{International Journal of Computer Assisted Radiology and Surgery}
\bvolume{19}(\bissue{1}),
\bfpage{1}--\blpage{9}
(\byear{2024})
\doiurl{10.1007/s11548-023-02942-x}
\end{barticle}
\endbibitem

\bibitem[\protect\citeauthoryear{Heiselman et~al.}{2020}]{heiselman2020intraoperative}
\begin{barticle}
\bauthor{\bsnm{Heiselman}, \binits{J.S.}},
\bauthor{\bsnm{Jarnagin}, \binits{W.R.}},
\bauthor{\bsnm{Miga}, \binits{M.I.}}:
\batitle{Intraoperative correction of liver deformation using sparse surface and vascular features via linearized iterative boundary reconstruction}.
\bjtitle{IEEE Transactions on Medical Imaging}
\bvolume{39}(\bissue{6}),
\bfpage{2223}--\blpage{2234}
(\byear{2020})
\doiurl{10.1109/TMI.2020.2967322}
\end{barticle}
\endbibitem

\bibitem[\protect\citeauthoryear{Clements et~al.}{2016}]{clements2016evaluation}
\begin{barticle}
\bauthor{\bsnm{Clements}, \binits{L.W.}},
\bauthor{\bsnm{Collins}, \binits{J.A.}},
\bauthor{\bsnm{Weis}, \binits{J.A.}},
\bauthor{\bsnm{Simpson}, \binits{A.L.}},
\bauthor{\bsnm{Adams}, \binits{L.B.}},
\bauthor{\bsnm{Jarnagin}, \binits{W.R.}},
\bauthor{\bsnm{Miga}, \binits{M.I.}}:
\batitle{Evaluation of model-based deformation correction in image-guided liver surgery via tracked intraoperative ultrasound}.
\bjtitle{Journal of Medical Imaging}
\bvolume{3}(\bissue{1}),
\bfpage{015003}--\blpage{015003}
(\byear{2016})
\doiurl{10.1117/1.JMI.3.1.015003}
\end{barticle}
\endbibitem

\bibitem[\protect\citeauthoryear{Yang et~al.}{2016}]{yang2016ultrasound}
\begin{barticle}
\bauthor{\bsnm{Yang}, \binits{M.}},
\bauthor{\bsnm{Ding}, \binits{H.}},
\bauthor{\bsnm{Zhu}, \binits{L.}},
\bauthor{\bsnm{Wang}, \binits{G.}}:
\batitle{Ultrasound fusion image error correction using subject-specific liver motion model and automatic image registration}.
\bjtitle{Computers in biology and medicine}
\bvolume{79},
\bfpage{99}--\blpage{109}
(\byear{2016})
\doiurl{10.1016/j.compbiomed.2016.06.025}
\end{barticle}
\endbibitem

\bibitem[\protect\citeauthoryear{Wang and Wang}{2025}]{wang2025eureg}
\begin{bchapter}
\bauthor{\bsnm{Wang}, \binits{H.}},
\bauthor{\bsnm{Wang}, \binits{Y.}}:
\bctitle{Eureg: End-to-end framework for efficient 2d-3d ultrasound registration}.
In: \bbtitle{International Conference on Medical Image Computing and Computer-Assisted Intervention},
pp. \bfpage{175}--\blpage{185}.
\bpublisher{Springer},
\blocation{Cham}
(\byear{2025}).
\doiurl{10.1007/978-3-032-04937-7_17}
\end{bchapter}
\endbibitem

\bibitem[\protect\citeauthoryear{Wei et~al.}{2019}]{Wei2019-xb}
\begin{bchapter}
\bauthor{\bsnm{Wei}, \binits{W.}},
\bauthor{\bsnm{Xu}, \binits{H.}},
\bauthor{\bsnm{Alpers}, \binits{J.}},
\bauthor{\bsnm{Tianbao}, \binits{Z.}},
\bauthor{\bsnm{Wang}, \binits{L.}},
\bauthor{\bsnm{Rak}, \binits{M.}},
\bauthor{\bsnm{Hansen}, \binits{C.}}:
\bctitle{Fast registration for liver motion compensation in ultrasound-guided navigation}.
In: \bbtitle{{IEEE} 16th International Symposium on Biomedical Imaging ({ISBI})}
(\byear{2019}).
\doiurl{10.1109/ISBI.2019.8759464}
\end{bchapter}
\endbibitem

\bibitem[\protect\citeauthoryear{Ronneberger et~al.}{2015}]{ronneberger2015u}
\begin{bchapter}
\bauthor{\bsnm{Ronneberger}, \binits{O.}},
\bauthor{\bsnm{Fischer}, \binits{P.}},
\bauthor{\bsnm{Brox}, \binits{T.}}:
\bctitle{U-net: Convolutional networks for biomedical image segmentation}.
In: \bbtitle{MICCAI},
pp. \bfpage{234}--\blpage{241}.
\bpublisher{Springer},
\blocation{Cham}
(\byear{2015}).
\doiurl{10.1007/978-3-319-24574-4_28}
\end{bchapter}
\endbibitem

\bibitem[\protect\citeauthoryear{Wu et~al.}{2023}]{Wu2023-ac}
\begin{barticle}
\bauthor{\bsnm{Wu}, \binits{M.}},
\bauthor{\bsnm{Qian}, \binits{Y.}},
\bauthor{\bsnm{Liao}, \binits{X.}},
\bauthor{\bsnm{Wang}, \binits{Q.}},
\bauthor{\bsnm{Heng}, \binits{P.-A.}}:
\batitle{Hepatic vessel segmentation based on {3D} swin-transformer with inductive biased multi-head self-attention}.
\bjtitle{BMC Med. Imaging}
\bvolume{23}(\bissue{1}),
\bfpage{91}
(\byear{2023})
\doiurl{10.1186/s12880-023-01045-y}
\end{barticle}
\endbibitem

\bibitem[\protect\citeauthoryear{Hille et~al.}{2024}]{Hille2024-sb}
\begin{barticle}
\bauthor{\bsnm{Hille}, \binits{G.}},
\bauthor{\bsnm{Jahangir}, \binits{T.}},
\bauthor{\bsnm{H{\"u}rtgen}, \binits{J.}},
\bauthor{\bsnm{Kreher}, \binits{R.}},
\bauthor{\bsnm{Saalfeld}, \binits{S.}}:
\batitle{Deep learning-based liver vessel segmentation}.
\bjtitle{Curr. Dir. Biomed. Eng.}
\bvolume{10}(\bissue{1}),
\bfpage{29}--\blpage{32}
(\byear{2024})
\doiurl{10.1515/cdbme-2024-0108}
\end{barticle}
\endbibitem

\bibitem[\protect\citeauthoryear{Velikova et~al.}{2023}]{velikova2023lotus}
\begin{bchapter}
\bauthor{\bsnm{Velikova}, \binits{Y.}},
\bauthor{\bsnm{Azampour}, \binits{M.F.}},
\bauthor{\bsnm{Simson}, \binits{W.}},
\bauthor{\bsnm{Gonzalez~Duque}, \binits{V.}},
\bauthor{\bsnm{Navab}, \binits{N.}}:
\bctitle{Lotus: learning to optimize task-based us representations}.
In: \bbtitle{International Conference on Medical Image Computing and Computer-Assisted Intervention},
pp. \bfpage{435}--\blpage{445}.
\bpublisher{Springer},
\blocation{Cham}
(\byear{2023}).
\doiurl{10.1007/978-3-031-43907-0_42}
\end{bchapter}
\endbibitem

\bibitem[\protect\citeauthoryear{Velikova et~al.}{2024}]{Velikova2024-fn}
\begin{barticle}
\bauthor{\bsnm{Velikova}, \binits{Y.}},
\bauthor{\bsnm{Simson}, \binits{W.}},
\bauthor{\bsnm{Azampour}, \binits{M.F.}},
\bauthor{\bsnm{Paprottka}, \binits{P.}},
\bauthor{\bsnm{Navab}, \binits{N.}}:
\batitle{{CACTUSS}: Common anatomical {CT-US} space for {US} examinations}.
\bjtitle{Int. J. Comput. Assist. Radiol. Surg.}
\bvolume{19}(\bissue{5}),
\bfpage{861}--\blpage{869}
(\byear{2024})
\doiurl{10.1007/s11548-024-03060-y}
\end{barticle}
\endbibitem

\bibitem[\protect\citeauthoryear{Lecomte-Denis et~al.}{2025}]{lecomte2025domain}
\begin{barticle}
\bauthor{\bsnm{Lecomte-Denis}, \binits{F.}},
\bauthor{\bsnm{Verde}, \binits{J.}},
\bauthor{\bsnm{Dillenseger}, \binits{J.-L.}},
\bauthor{\bsnm{Cotin}, \binits{S.}}:
\batitle{Domain agnostic 2d-3d deformable registration application to fluoroscopic guidance without contrast agent}.
\bjtitle{Medical Image Analysis}
\bvolume{105},
\bfpage{103688}
(\byear{2025})
\doiurl{10.1016/j.media.2025.103688}
\end{barticle}
\endbibitem

\bibitem[\protect\citeauthoryear{Park et~al.}{2020}]{park2020contrastive}
\begin{bchapter}
\bauthor{\bsnm{Park}, \binits{T.}},
\bauthor{\bsnm{Efros}, \binits{A.A.}},
\bauthor{\bsnm{Zhang}, \binits{R.}},
\bauthor{\bsnm{Zhu}, \binits{J.-Y.}}:
\bctitle{Contrastive learning for unpaired image-to-image translation}.
In: \bbtitle{Computer Vision--ECCV 2020: 16th European Conference, Glasgow, UK, August 23--28, 2020, Proceedings, Part IX 16},
pp. \bfpage{319}--\blpage{345}.
\bpublisher{Springer},
\blocation{Cham}
(\byear{2020}).
\doiurl{10.1007/978-3-030-58545-7_19}
\end{bchapter}
\endbibitem

\bibitem[\protect\citeauthoryear{Wasserthal et~al.}{2023}]{wasserthal2023totalsegmentator}
\begin{barticle}
\bauthor{\bsnm{Wasserthal}, \binits{J.}},
\bauthor{\bsnm{Breit}, \binits{H.-C.}},
\bauthor{\bsnm{Meyer}, \binits{M.T.}},
\bauthor{\bsnm{Pradella}, \binits{M.}},
\bauthor{\bsnm{Hinck}, \binits{D.}},
\bauthor{\bsnm{Sauter}, \binits{A.W.}},
\bauthor{\bsnm{Heye}, \binits{T.}},
\bauthor{\bsnm{Boll}, \binits{D.T.}},
\bauthor{\bsnm{Cyriac}, \binits{J.}},
\bauthor{\bsnm{Yang}, \binits{S.}},
\bauthor{\bsnm{Baumgartner}, \binits{C.F.}}:
\batitle{Totalsegmentator: robust segmentation of 104 anatomic structures in ct images}.
\bjtitle{Radiology: Artificial Intelligence}
\bvolume{5}(\bissue{5}),
\bfpage{230024}
(\byear{2023})
\doiurl{10.1148/ryai.230024}
\end{barticle}
\endbibitem

\bibitem[\protect\citeauthoryear{Oktay et~al.}{2018}]{Oktay2018AttentionUNet}
\begin{bchapter}
\bauthor{\bsnm{Oktay}, \binits{O.}},
\bauthor{\bsnm{Schlemper}, \binits{J.}},
\bauthor{\bsnm{Le~Folgoc}, \binits{L.}},
\bauthor{\bsnm{Lee}, \binits{M.}},
\bauthor{\bsnm{Heinrich}, \binits{M.P.}},
\bauthor{\bsnm{Misawa}, \binits{K.}},
\bauthor{\bsnm{Mori}, \binits{K.}},
\bauthor{\bsnm{McDonagh}, \binits{S.G.}},
\bauthor{\bsnm{Hammerla}, \binits{N.Y.}},
\bauthor{\bsnm{Kainz}, \binits{B.}},
\bauthor{\bsnm{Glocker}, \binits{B.}},
\bauthor{\bsnm{Rueckert}, \binits{D.}}:
\bctitle{Attention u-net: Learning where to look for the pancreas}.
In: \bbtitle{Medical Imaging with Deep Learning (MIDL)}
(\byear{2018}).
\bcomment{OpenReview: \url{https://openreview.net/forum?id=Skft7cijM}}
\end{bchapter}
\endbibitem

\end{thebibliography}

\end{document}